\documentclass{nst}

\usepackage{subfigure,dcolumn}
\usepackage{graphicx}
\usepackage{float}
\usepackage{amsmath, amssymb} 
\usepackage{multirow}
\usepackage{makecell}
\usepackage{booktabs}
\usepackage{array}
\usepackage[T2A,T1]{fontenc}
\usepackage[russian,english]{babel}

\usepackage{listings}
\begin{document}

\title [mode = title]{Machine Learning methods for event classification and vertex reconstruction of the ${}^{12}\mathrm{C} + {}^{12}\mathrm{C}$ reaction with the MATE-TPC}       
\thanks{This work is supported by the National Natural Science Foundation of China (Grant Nos. 12475120, 12435010, 11490560, 12505146, 12335009) and the National Key Research and Development Program of China (Grant No. 2022YFA1602301, 2022YFA1602304).}

\author{Minghui Zhang}
\affiliation{College of Science, Southern University of Science and Technology, Shenzhen, 518055, China}
\author{Xiaobin Li}
\thanks{Co-first authors.}
\affiliation{College of Science, Southern University of Science and Technology, Shenzhen, 518055, China}
\author{Jie Chen}
\email[Corresponding author, ]{chenjie@sustech.edu.cn}
\affiliation{College of Science, Southern University of Science and Technology, Shenzhen, 518055, China}
\author{Ningtao Zhang}
\email[Corresponding author, ]{zhangningtao@impcas.ac.cn}
\affiliation{Institute of Modern Physics, Chinese Academy of Sciences, Lanzhou 730000, China}
\affiliation{School of Nuclear Science and Technology, University of Chinese Academy of Sciences, Beijing 100049, China}
\author{Fenhua Lu}
\affiliation{College of Science, Southern University of Science and Technology, Shenzhen, 518055, China}
\author{Junrui Ma}
\affiliation{College of Science, Southern University of Science and Technology, Shenzhen, 518055, China}
\author{Jiazhen Yan}
\affiliation{School of Physics, Peking University, Beijing 100871, China}
\author{Wanqin Tu}
\affiliation{College of Science, Southern University of Science and Technology, Shenzhen, 518055, China}
\author{Xiaodong Tang}
\affiliation{Institute of Modern Physics, Chinese Academy of Sciences, Lanzhou 730000, China}
\affiliation{School of Nuclear Science and Technology, University of Chinese Academy of Sciences, Beijing 100049, China}
\author{Bingshui Gao}
\affiliation{Institute of Modern Physics, Chinese Academy of Sciences, Lanzhou 730000, China}
\affiliation{School of Nuclear Science and Technology, University of Chinese Academy of Sciences, Beijing 100049, China}
\author{Chengui Lu}
\affiliation{Institute of Modern Physics, Chinese Academy of Sciences, Lanzhou 730000, China}
\affiliation{School of Nuclear Science and Technology, University of Chinese Academy of Sciences, Beijing 100049, China}
\author{Zhichao Zhang}
\affiliation{Institute of Modern Physics, Chinese Academy of Sciences, Lanzhou 730000, China}
\affiliation{School of Nuclear Science and Technology, University of Chinese Academy of Sciences, Beijing 100049, China}
\author{Jinlong Zhang}
\affiliation{Institute of Modern Physics, Chinese Academy of Sciences, Lanzhou 730000, China}
\affiliation{School of Nuclear Science and Technology, University of Chinese Academy of Sciences, Beijing 100049, China}
\author{Weiping Liu}
\email[Corresponding author, ]{liuwp@sustech.edu.cn}
\affiliation{College of Science, Southern University of Science and Technology, Shenzhen, 518055, China}

\begin{abstract}

In modern nuclear physics experiments, identifying the events of interest is challenging for nuclear reaction studies with the active target Time Projection Chamber (TPC). In this work, machine learning techniques are employed to analyze the complex data of ${}^{12}\mathrm{C}+{}^{12}\mathrm{C}$ fusion reaction from a TPC named MATE (multi-purpose active-target time projection chamber for nuclear experiments). Specifically, we successfully applied
 Residual Neural Network (ResNet-50, ResNet-34 and ResNet-18) and Visual Geometry Group (VGG-19) to classify elastic scattering and fusion reaction events from the ${}^{12}\mathrm{C}+{}^{12}\mathrm{C}$ reaction. The classification results of the four models are nearly identical, with accuracies of approximately $97\%$ for the simulated data and $90\%$ for the experimental data. Moreover, these approaches successfully identify some events that are misclassified by traditional methods. These models are also applied to classify events from different fusion reaction channels, their classification accuracies are approximately $95\%$ on the simulated data. In addition, a Convolutional Neural Network (CNN) model is developed to reconstruct the reaction vertex, providing an alternative strategy for vertex reconstruction. These results indicate that machine learning techniques can effectively classify the reaction events of different reaction channels and reconstruct the reaction vertex, thereby paving the way for future analyses of complex nuclear reaction data.

\end{abstract}

\keywords{MATE-TPC, machine learning, event classification, vertex reconstruction, Residual Neural Network}

\maketitle

\section{Introduction}

Nuclear reactions with radioactive beams play an important role in enhancing our understanding of nuclear force and reaction dynamics. Meanwhile, they also provide essential nuclear data for nuclear engineering and astrophysics. With continuous progress in radioactive beams facilities, the quality of radioactive beams has markedly improved in recent years \cite{Blumenfeld2013,Stracener2003, Bricault1997}. However, compared with stable beams, the intensity of radioactive beams is still lower by at least 4 orders of magnitude, which poses a great challenge for experimental studies, especially for beams of nuclei far from stability. The active target time projection chamber is one of the most powerful experimental equipments for nuclear reaction studies with radioactive beams, owing to its precise three-dimensional track reconstruction capability, particle identification performance, and ability to enhance experimental luminosity. Therefore, it has undergone rapid development over the past two decades~\cite{bazin2020low, ayyad2018physics}. Moreover, it offers additional advantages, including the nearly 4$\pi$ solid-angle coverage, high spatial resolution, and enhanced effective target thickness, without inducing additional significant losses in energy resolution. These characteristics make the active target time projection chamber particularly well-suited for experiments with radioactive beams and rare event detection. Currently, the active target time projection chamber is widely used in various nuclear experiments for the detection of low-energy charged particles. Recently developed active target time projection chambers include the AT-TPC at FRIB \cite{MITTIG2015494}, ACTAR-TPC at GANIL \cite{GIOVINAZZO2020}, TexAT at Texas A\&M \cite{KOSHCHIY2020}, O-TPC at Warsaw \cite{OBERLA2016}, fMata-TPC at Fudan University \cite{WU2023168528}, MATE-TPC at IMP \cite{ZHANG2021165740, Li2024}, and MTPC at CSNS \cite{LI2024}. These devices have been applied to various nuclear reaction studies \cite{Chen2025,Chen2024, Zhang2023prc}.

The identification of events of interest is important for the measurement of the reactions with low cross sections. However, the data analysis of the active target time projection chamber is critically challenging due to the large data volume, complex three-dimensional track topology, and the presence of background noise. Currently, traditional data processing methods often rely on complicated tracking and event selection, which typically require specialized discrimination methods or multi-step processing for each type of event. These approaches often demand substantial time and computational resources, rendering them increasingly inadequate for research with the next-generation experimental facilities or for complex nuclear reaction processes. Recent advancements in machine learning present a promising alternative approach. Machine learning techniques exhibit efficient data processing capabilities, leveraging powerful feature extraction to handle large-scale and high-dimensional datasets. This includes unstructured data (such as images, texts, etc.) that are challenging for traditional methods. 

Recent studies have demonstrated that machine learning techniques can enhance the accuracy and efficiency of nuclear reaction data analysis \cite{RevModPhys.94.031003, he2023high, GAO2021, YUAN2024, SHANG2022, HE2021, KIM2023,QIAN2021,MAYER2021,LI2022,DELAQUIS2018}. To begin with, supervised learning methods have been widely applied. For example, Kuchera et al. employed CNN to classify proton scattering events in AT-TPC, achieving precision values exceeding 0.95 on both experimental and simulated datasets \cite{KUCHERA2019156}. Similarly, Wu et al. used two CNNs, ResNet and VGG-16, to identify ${}^{12}\mathrm{C}$ events from background events in simulated data, achieving a precision above 0.93 \cite{WU2023168528}. In addition to event classification, unsupervised learning methods were applied to the latent space of deep learning models to cluster AT-TPC events \cite{SOLLI2021165461}. Other machine learning models have also proven effective. The random forest algorithm was used to separate inelastic scattering events on carbon and oxygen contaminants from ${}^{96}\mathrm{Mo}(p, p')$ data \cite{GHIMIRE2025165649}. Meanwhile, the PointNet architecture was used to classify fission and non-fission events from AT-TPC, achieving an accuracy of about 0.99 \cite{DEY2025170002}. Although machine learning techniques have achieved significant progress in nuclear reaction event classification, most existing studies have primarily focused on reaction channels with clear physical signatures and distinctive topological features, such as 3$\alpha$ decays or scattering events without fusion contamination. In contrast, the effective discrimination between events, that exhibit highly similar topologies and energy-deposition patterns, has not yet been systematically investigated. Moreover, the application of machine learning to reaction-vertex reconstruction in heavy-ion experiments remains relatively limited, particularly with realistic experimental data. Therefore, systematic studies are necessary that simultaneously address complex event classification and high-precision reaction-vertex reconstruction.

In this study, we analyze the experimental data obtained from the 1024-channel MATE-TPC, which aims at measuring the fusion reaction cross section of ${}^{12}\mathrm{C} + {}^{12}\mathrm{C}$ near the Coulomb barrier. The dataset contains a large number of elastic scattering and fusion reaction events with highly similar physical characteristics, which significantly complicates the determination of the fusion reaction cross section. Therefore, it is crucial to identify elastic scattering and fusion reaction events. Owing to the advantages of ResNet and VGG in image recognition, we employ the ResNet-50, ResNet-34, ResNet-18, and VGG-19 architectures to classify elastic scattering and fusion reaction events from experimental and simulated data. Through analysis of misclassified events, we find that machine learning can correctly label some events that are misclassified by traditional methods, which improves the accuracy of data analysis. In addition, we extend the classification task to classify the events for different fusion reaction channels, enabling more accurate selection of events of interest. Furthermore, the precise reconstruction of the reaction vertex in heavy-ion experiments is particularly challenging due to the small exit angles and short track ranges of the reaction products. In this work, a CNN model is developed to effectively reconstruct the reaction vertex. 

The structure of this article is as follows. In Section~\ref{sec2}, we give a simple introduction on the experimental measurement of fusion reaction cross section of ${}^{12}\mathrm{C} + {}^{12}\mathrm{C}$ near the Coulomb barrier. Section~\ref{sec3} describes the data processing methods, while Section~\ref{sectr} introduces the traditional data analysis methods. Section~\ref{sec4} discusses the use of MATEROOT \cite{MATEroot2025} to simulate nuclear reaction events generated by MATE-TPC through Monte Carlo methods. Section~\ref{sec5} focuses on the application of machine learning to event classification and reaction vertex reconstruction, including the use of ResNet and VGG models for various classification tasks and a CNN for vertex reconstruction. Section~\ref{sec11} provides a summary and discussion of the results.

\section{Experimental details}\label{sec2}
\begin{figure}[htbp]
    \centering

    \setlength{\abovecaptionskip}{0pt}  
    \setlength{\belowcaptionskip}{0pt}  
    \setlength{\subfigbottomskip}{-10pt}  
    \setlength{\subfigcapskip}{0pt}

    \includegraphics[width=0.45\textwidth]{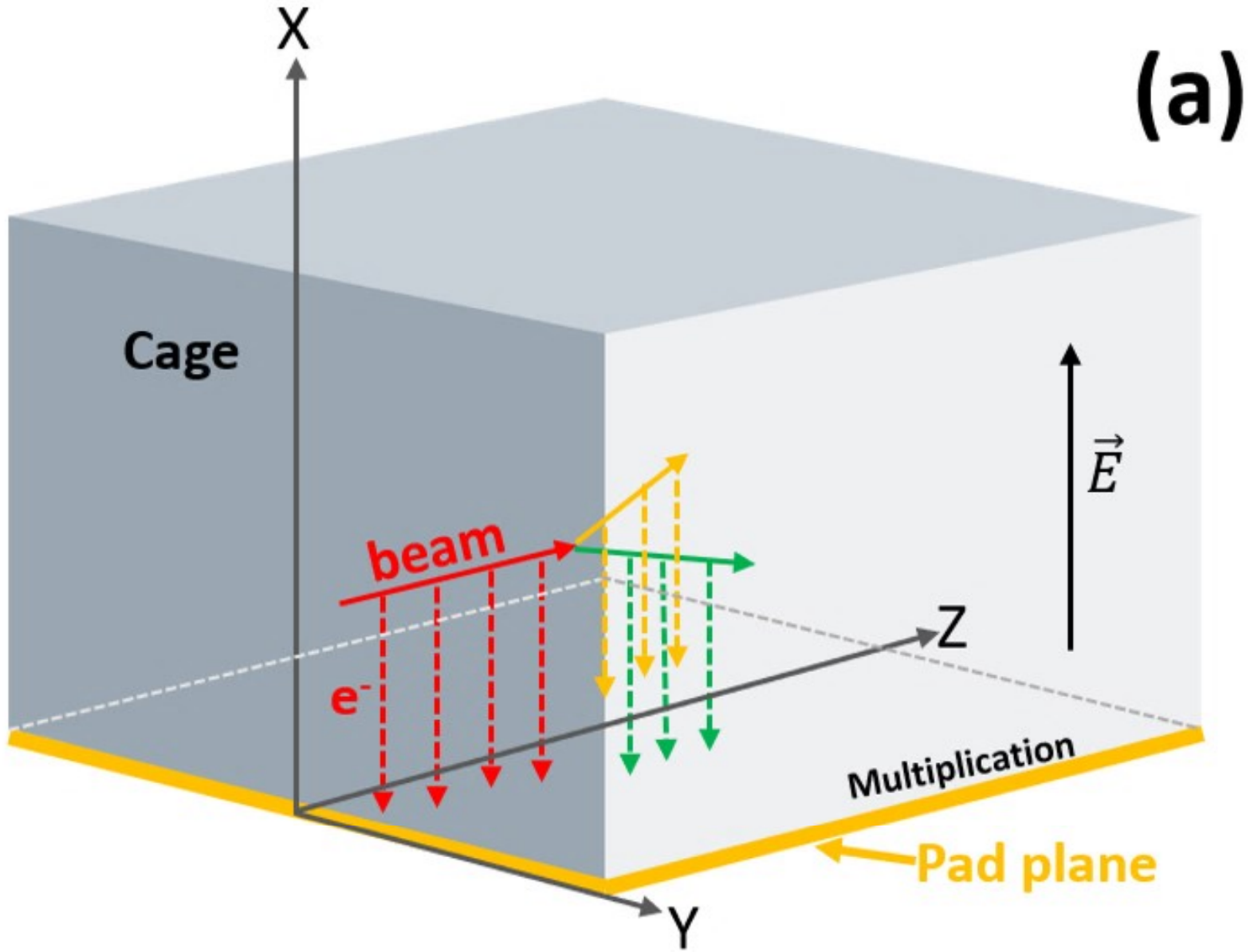}\\[1pt]
    \includegraphics[width=0.45\textwidth]{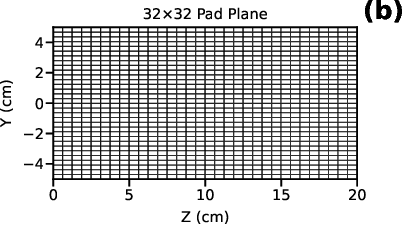}

    \caption{Sketches showing the working principle of the MATE-TPC (a) and the layout of the pad plane (b).
}
    \label{fig1}
\end{figure}

 MATE-TPC consists of three major parts: drift cage, electron multiplication unit and pad plane. The working principle is illustrated in Fig. \ref{fig1}(a). The detector has an active volume of $100~\mathrm{mm} \times 200~\mathrm{mm} \times 200~\mathrm{mm~(H)}$. And the pad plane of MATE-TPC, with a total area of \(100 \times 200~\mathrm{mm}^2\), is segmented into \(32 \times 32\) rectangular pads of \(3 \times 6~\mathrm{mm}^2\), as shown in Fig.~\ref{fig1}(b). The detailed description and performance of the MATE-TPC can be found in Ref.~\cite{ZHANG2021165740}.

\begin{figure}[htbp]
    \centering
    \includegraphics[width=0.45\textwidth]{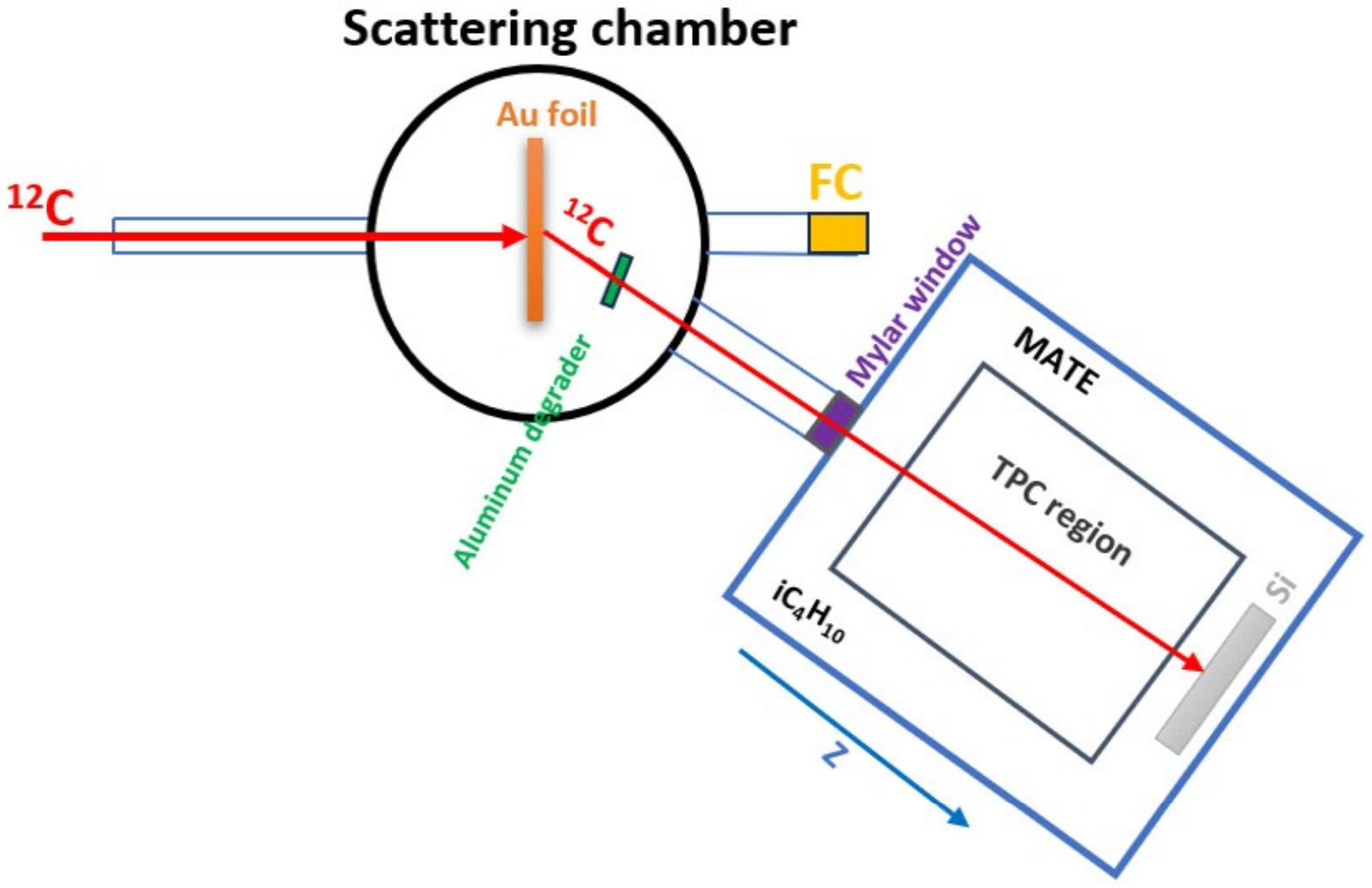}  
    \caption{Sketch of the experimental setup.}
    \label{fig2}
\end{figure}

The data used in this work are from the measurement of the ${}^{12}\mathrm{C} + {}^{12}\mathrm{C}$ fusion reaction cross section around the Coulomb barrier, performed with the 1024-channel MATE-TPC at a terminal of the Sector Focused Cyclotron (SFC) in the Heavy Ion Research Facility in Lanzhou (HIRFL) \cite{XIA200211}. The goal of this measurement is to study the reaction mechanism of ${}^{12}\mathrm{C} + {}^{12}\mathrm{C}$ fusion reaction, which is important during the massive star evolution and the ignition phase of Type Ia supernovae and Superbursts \cite{Nan2024, ZHANG2021165740, tang201912c+, RevModPhys.86.317,PhysRevC.76.035802, Wang20251, Tang20191, Liu20241}. The present work focuses on a center-of-mass energy range of $E_{\mathrm{cm}} = 8.9$--$21~\mathrm{MeV}$, within which variations in beam
energy have only a limited impact on track morphology, and during which the reaction cross sections and dominant reaction channels remained essentially unchanged. The major channels of the ${}^{12}\mathrm{C} + {}^{12}\mathrm{C}$ fusion reaction are as follows \cite{Zhang2021}:
\begin{align}
{}^{12}\mathrm{C} + {}^{12}\mathrm{C} 
&\rightarrow {}^{24}\mathrm{Mg} 
\rightarrow {}^{23}\mathrm{Na} + p \notag \\
&\hspace{1.41cm}\rightarrow {}^{20}\mathrm{Ne} + \alpha \notag \\
&\hspace{1.41cm}\rightarrow {}^{23}\mathrm{Mg} + n \notag \\
&\hspace{1.41cm}\rightarrow {}^{16}\mathrm{O} + 2\alpha \notag \\
&\hspace{1.41cm}\rightarrow {}^{16}\mathrm{O} + {}^{8}\mathrm{Be} 
\label{eq1}
\end{align}
Among them, ${}^{12}\mathrm{C}({}^{12}\mathrm{C},p){}^{23}\mathrm{Na}$ and ${}^{12}\mathrm{C}({}^{12}\mathrm{C},\alpha){}^{20}\mathrm{Ne}$ have high reaction cross sections and are the dominant reaction channels below the Coulomb barrier, making them the primary targets of the experiment. Since a more detailed description of the experiment has been given in the previous publication \cite{Wang_2022}, we only summarize the experimental setup here. The ${}^{12}\mathrm{C}^{4+}$ primary beam is emitted from SFC with a beam intensity in the microampere range ($10^{12}$ pps), which cannot be directly injected into the MATE-TPC. In order to obtain a low intensity beam of the order of $10^2$ cps, a gold foil with a thickness of 0.81 $\mathrm{mg/cm^2}$ was used in the scattering target chamber to scatter the ${}^{12}\mathrm{C}$ beam, as shown in Fig. \ref{fig2}. The calibrated beam was scattered from a gold foil and subsequently passed through a 28.3~$\mu$m aluminum degrader. MATE-TPC was installed at a scattering angle of $30^\circ$, and separated from the vacuum target chamber through a Mylar window with a thickness of 10 $\mu\mathrm{m}$. The scattered ${}^{12}\mathrm{C}$ beam entered the MATE-TPC through the Mylar window, passed through a 70 mm gas dead region, and then entered the sensitive region with a total length of 200 mm along the Z direction of MATE-TPC. The gas target of the MATE-TPC is pure isobutane at a pressure of 50 mbar. Reactions occurred between the carbon beam and the isobutane gas target and were detected by the MATE-TPC. The experimental setup is shown in Fig. \ref{fig2}.

\section{Data processing}\label{sec3}
\begin{figure*}[htbp]
    \centering

    \setlength{\abovecaptionskip}{2pt}  
    \setlength{\belowcaptionskip}{0pt}  
    \setlength{\subfigcapskip}{0pt}

    \subfigure[]{
        \includegraphics[width=0.45\textwidth]{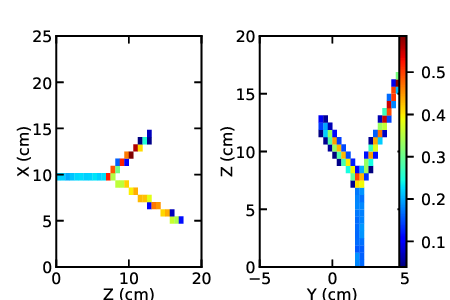}
    }
    \subfigure[]{
        \includegraphics[width=0.45\textwidth]{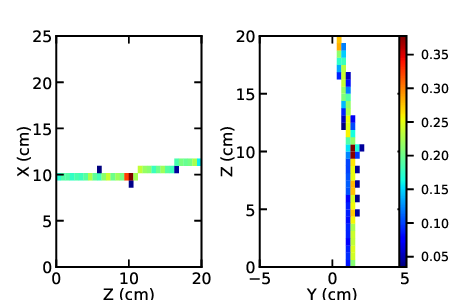}
    }\\[-0.2cm]
    \subfigure[]{
        \includegraphics[width=0.45\textwidth]{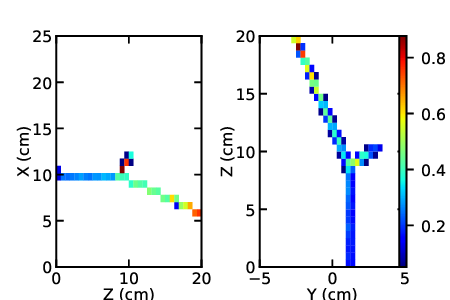}
    }
    \subfigure[]{
        \includegraphics[width=0.45\textwidth]{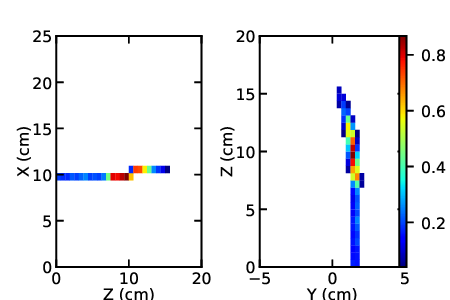}
    }\\[-0.2cm]
    \subfigure[]{
        \includegraphics[width=0.45\textwidth]{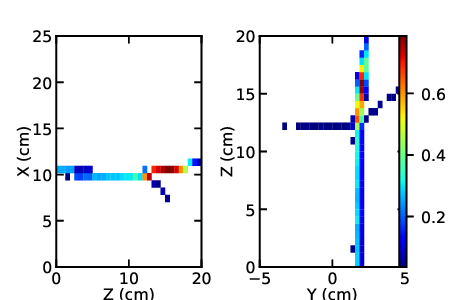}
    }
    \subfigure[]{
        \includegraphics[width=0.45\textwidth]{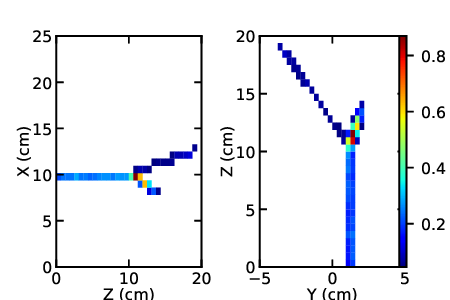}
    }

    \caption{Three typical elastic scattering events (a-c) and fusion reaction events (d-f) observed in the experiment. Each figure shows the projections of the three-dimensional trajectories of the beam and reaction products onto two different planes in the MATE-TPC. The positive Z direction is defined as the beam direction, while the negative X direction is the drift direction toward the pad plane. The intersection point of the tracks corresponds to the reaction vertex. The color scale represents the deposited charge in artifical unit. }
    \label{fig3}
\end{figure*}

 Events of interest are selected according to the track features of all detected charged particles, thereby enabling the measurement of the reaction cross section. Due to the influence of Coulomb barrier, the number of elastic (or Rutherford) scattering events in the low-energy region exceeds that of fusion reaction events. Therefore, this poses a challenge for the selection of fusion reaction events. Fig. \ref{fig3} shows the three-dimensional tracks of two types of events projected onto two different planes (Z-X plane and Y-Z plane) within the MATE-TPC. The color scale represents the deposited charge. For elastic scattering events, the track projections typically exhibit a "Y"-shaped (as shown in Fig. \ref{fig3}(a) and (c)) or single-track pattern (as shown in Fig. \ref{fig3}(b)). However, in certain cases, fusion reaction events may also display similar morphological features in their track projections. For instance, Fig. \ref{fig3}(f) exhibits a "Y"-shaped pattern, while Fig. \ref{fig3}(d) shows a single-track pattern. These similarities cause significant interference in the determination of the fusion reaction cross section. Thus, a major task in data analysis is to distinguish fusion reaction events and elastic scattering events. In addition, it is essential to reconstruct the reaction vertex, which is critical for determining the energy of the reaction.

\subsection{Traditional data analysis methods}\label{sectr}
 The Hough transform is used to select particle tracks and reconstruct the reaction vertex by extrapolating tracks to the beam axis. For proton emission channels, since the energy loss in the sensitive region of the detector is too small or absent in the experimental setup, MATE-TPC is unable to provide track information. Therefore, the reaction vertex for such events can only be inferred from the point at which a sudden change in energy loss is observed. In addition, we classify and identify the reaction type based on the number of reaction particle tracks N recorded by the MATE-TPC. The specific steps are as follows: (1) N=1: If the angle change of the reaction product in the sensitive zone is small and the energy loss curve is close to the beam particle, the event is classified as a Rutherford scattering event. Otherwise, it is identified as a fusion event corresponding to proton or neutron emission channels. (2) N=2: If the energy loss of two tracks is close to ${}^{12}\mathrm{C}$ particle and the angle between them is around $90^\circ$, the event is determined to be an elastic scattering event; If the energy loss of two tracks is significantly different, and one of them matches the energy loss characteristics of $\alpha$ particle, the event is classified as a fusion reaction. (3) $\mathrm{N} \geq 3$: Such events do not conform to the characteristics of scattering events and are directly judged as fusion events. By applying this analysis process, approximately 4000 fusion reaction events and 2000 elastic scattering events were identified from the experimental data.

\subsection{Simulation of elastic scattering and fusion reaction events}\label{sec4}
\begin{figure*}[htbp]
    \centering

    \setlength{\abovecaptionskip}{2pt}  
    \setlength{\belowcaptionskip}{0pt}  
    \setlength{\subfigbottomskip}{5pt}  
    \setlength{\subfigcapskip}{0pt}

    \subfigure[]{
        \includegraphics[width=0.45\textwidth]{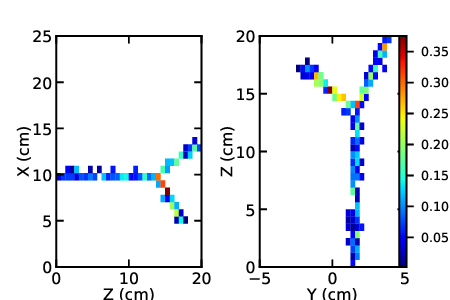}
    }
    \subfigure[]{
        \includegraphics[width=0.45\textwidth]{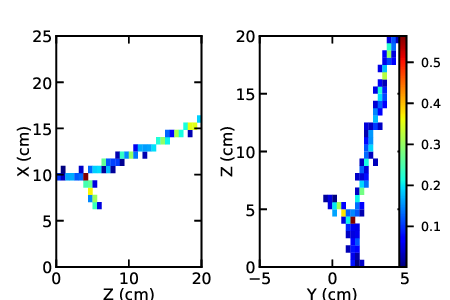}
    }\\[-0.2cm]
    
    \subfigure[]{
        \includegraphics[width=0.45\textwidth]{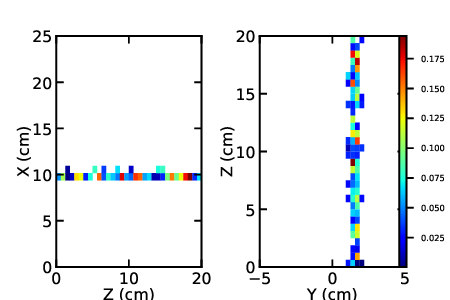}
    }
    \subfigure[]{
        \includegraphics[width=0.45\textwidth]{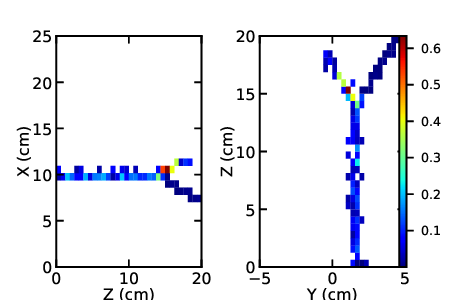}
    }\\[-0.2cm]
    \subfigure[]{
        \includegraphics[width=0.45\textwidth]{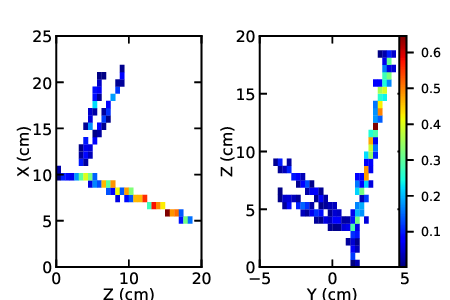}
    }
    \subfigure[]{
        \includegraphics[width=0.45\textwidth]{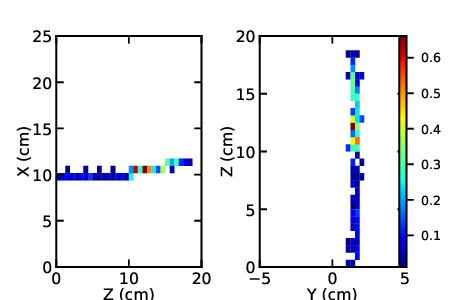}
    }

    \caption{Three typical simulated elastic scattering events (a-c) and fusion reaction events corresponding to the ${}^{12}\mathrm{C}({}^{12}\mathrm{C},\alpha){}^{20}\mathrm{Ne}$ (d), ${}^{12}\mathrm{C}({}^{12}\mathrm{C},2\alpha){}^{16}\mathrm{O}$ (e), and ${}^{12}\mathrm{C}({}^{12}\mathrm{C},n){}^{23}\mathrm{Mg}$ (f) channels.}
    \label{fig4}
\end{figure*}

Training neural network models with experimental data requires prior classification. However, traditional data processing methods may cause false positives, resulting in incorrect labels. In addition, the limited amount of experimental data often fails to meet the needs of machine learning. Therefore, data simulation is essential to ensure sufficient quantity and quality of training data. We utilize the data analysis platform MATEROOT, which is developed based on Geant4 and the ROOT framework, to simulate nuclear reaction events produced by the MATE-TPC through Monte Carlo methods. The simulation program is based on energy deposition information generated by Geant4, and the physical hit events are progressively converted into electronic output signals consistent with the experimental data through a series of digitization and signal reconstruction modules. First, the three-dimensional geometry of the MATE-TPC detector and the pad layout are defined using dedicated geometry and channel mapping files. Subsequently, during the signal generation stage, the production of primary electrons, the avalanche multiplication process, and their drift and collection in the electric field are simulated, taking into account factors such as electronic gain and energy threshold. Next, the resulting pulse signals are processed through peak analysis and threshold filtering to obtain spatial hit points. The Hough transform is then used to select particle tracks, and RANSAC together with clustering algorithms is employed to identify and fit multi-track events. This framework includes the geometric configuration, electronic response, reconstruction algorithms, and physical parameterization, establishing a full-chain digitization workflow for generating realistic simulated data. By employing MATEROOT, we simulate 48,000 elastic scattering events corresponding to the reaction channel ${}^{12}\mathrm{C}({}^{12}\mathrm{C},{}^{12}\mathrm{C}){}^{12}\mathrm{C}$. For fusion reaction channels, we simulate 13,000 events for each of the following reactions: ${}^{12}\mathrm{C}({}^{12}\mathrm{C},p){}^{23}\mathrm{Na}$, ${}^{12}\mathrm{C}({}^{12}\mathrm{C},\alpha){}^{20}\mathrm{Ne}$, ${}^{12}\mathrm{C}({}^{12}\mathrm{C},n){}^{23}\mathrm{Mg}$, ${}^{12}\mathrm{C}({}^{12}\mathrm{C},{}^{8}\mathrm{Be}){}^{16}\mathrm{O}$ and ${}^{12}\mathrm{C}({}^{12}\mathrm{C},2\alpha){}^{16}\mathrm{O}$, resulting in a total of 65,000 simulated fusion events. The angular distributions of all simulated events are assumed to be isotropic in the center-of-mass frame. All simulation parameters are consistent with those described in Section~\ref{sec2}. As shown in Fig. \ref{fig4}(a)-(c), there are three typical simulated elastic scattering events; Fig. \ref{fig4}(d)-(f) are three typical simulated fusion reaction events.

\subsection{CNN for data processing}\label{sec5}
ResNet \cite{Koonce2021,Liang2020, Li2018} and VGG \cite{Sengupta2019,Tammina2019} are two representative CNN architectures in the field of computer vision and have been widely applied to various image processing tasks. ResNet is mainly used to solve the problems of vanishing and exploding gradient encountered during the training process. It is a breakthrough model in the ImageNet competition, particularly outstanding in image classification tasks. ResNet includes convolutional layers, batch normalization layers, ReLU activation functions, residual blocks, pooling layers and fully connected layers. VGG is applied to large-scale image classification tasks, such as the ImageNet Large Scale Visual Recognition Challenge (ILSVRC), and it mainly enhances image feature extraction capabilities and classification accuracy by deepening the network architecture and employing small convolutional kernels. VGG includes multiple consecutive $3\times3$ convolutional layers, ReLU activations, $2\times2$ pooling layers, fully connected layers, and a final Softmax layer. 
 In this study, we employ ResNet-50, ResNet-34, ResNet-18, and VGG-19 to process nuclear reaction events recorded by the MATE-TPC.  Specifically, we use the four models for multiple tasks: (1) classification of elastic scattering and fusion reaction events generated from the ${}^{12}\mathrm{C} + {}^{12}\mathrm{C}$ reaction; (2) classification of events originating from different fusion reaction channels. The data are shown in Fig. \ref{fig4}. Each image is scaled to $224\times224$ pixels and represents the projections of the three-dimensional particle tracks on two planes.

\subsubsection{Classification of elastic scattering and fusion reaction events}\label{sec7}
The classification of elastic scattering and fusion reaction events is performed using ResNet-50, ResNet-34, ResNet-18, and VGG-19. First, we train these models on simulated data for the classification of elastic scattering and fusion reaction events. The Cross Entropy Loss function is used to measure the difference between the predicted probability distribution and the true class distribution. Model parameters are optimized using the Adam (Adaptive Moment Estimation) optimizer, with a learning rate set to 0.0001 and a batch size of 64. To address uncertainties, the models are trained multiple times on simulated data using different training-to-testing split ratios (8:2, 7:3, and 6:4). The performance of the models trained on simulated data is evaluated with experimental data.

The experimental data consist of 3773 fusion reaction events and 1621 elastic scattering events. The labels of the experimental data are obtained through traditional analysis methods. Concerning experimental uncertainty, the experimental data are preprocessed using conventional physics-based selection criteria to remove background and nonphysical events, retaining only elastic scattering and fusion reaction events. Consequently, the impact of experimental background on the classification results is almost completely eliminated. The classification results of the four models are compared in Table \ref{tab2}. It can be observed that the classification results of the four models are nearly identical, with accuracies of approximately $97\%$ for the simulated data and $90\%$ for the experimental data. 
The results show that variations in the training/testing split have only a minor effect on the classification performance, with stable accuracies obtained for both simulated and experimental data, thereby confirming the statistical stability and reproducibility of the results.

\begin{figure}[htbp]
    \raggedright
    \setlength{\abovecaptionskip}{1pt}  
    \setlength{\belowcaptionskip}{0pt}  
    \setlength{\subfigbottomskip}{0pt}  
    \setlength{\subfigcapskip}{0pt}

    \includegraphics[width=0.45\textwidth]{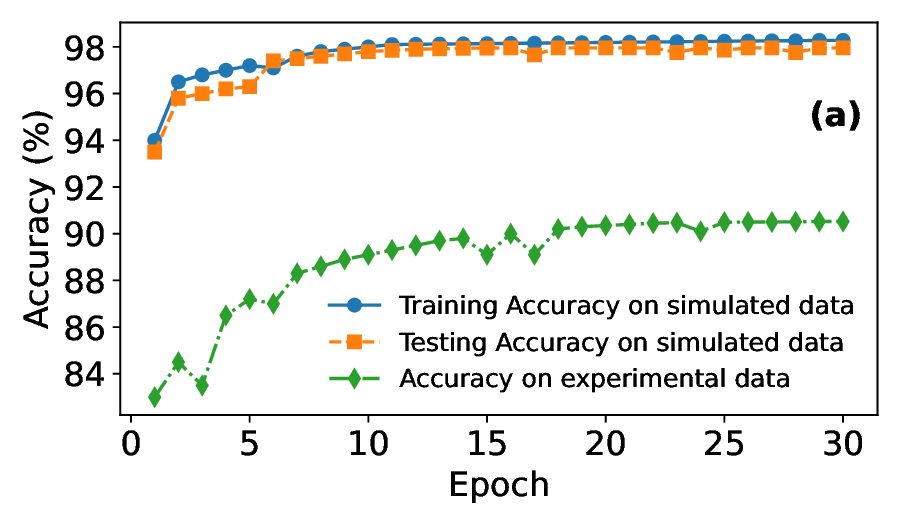}\\[-3pt]
    \includegraphics[width=0.45\textwidth]{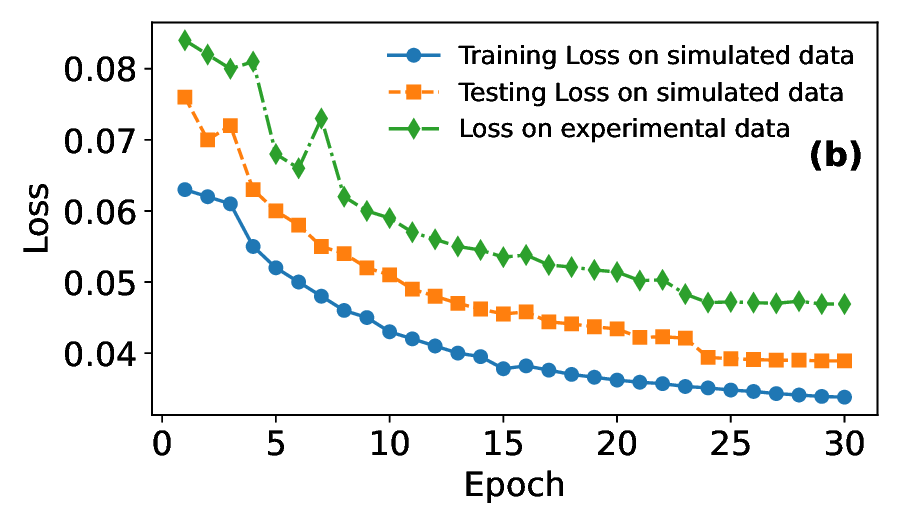}\\[-3pt]
    \includegraphics[width=8.5cm]{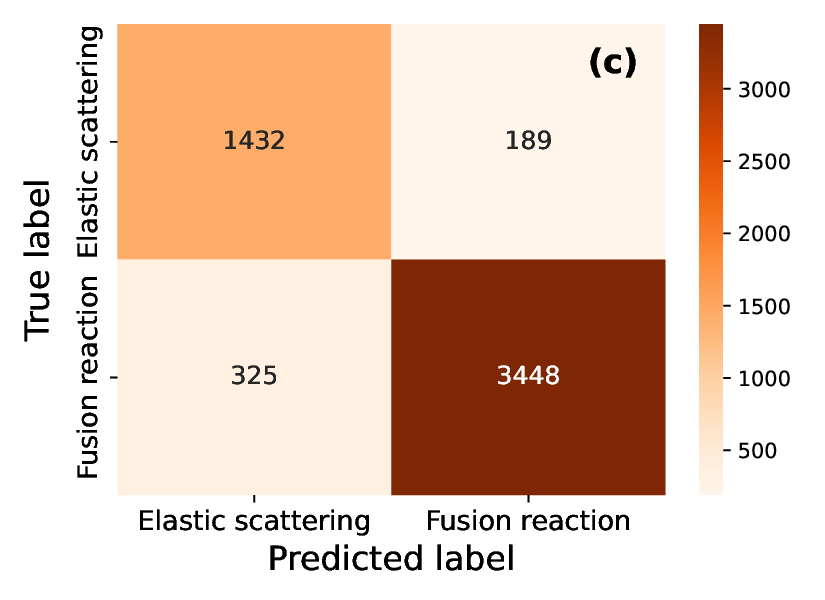}  

    \caption{The accuracy (a) and loss (b) of the training and testing sets over epoch for event classification using ResNet-50. (c) Confusion matrix for classifying events in the experimental data. The values along the diagonal from the top left to the bottom right represent the number of correctly classified events for each reaction type, while the values along the diagonal from the top right to the bottom left indicate the number of misclassified events between the two types.}
    \label{fig5}
\end{figure}

 To check for overfitting during the training process, the learning curves are plotted. Figs. \ref{fig5} (a) and (b) show the accuracy and loss of the training and testing sets over epoch, respectively. The training and testing curves exhibit consistent trends without significant divergence, indicating that no overfitting occurred during the training process. Fig. \ref{fig5} (c) shows the confusion matrix for classifying elastic scattering events and fusion reaction events in the experimental data using ResNet-50.



\begin{table}[htbp]
\centering
\caption{Classification results of four CNN models for elastic scattering and fusion reaction events on experimental and simulated data.}
\label{tab2}
\renewcommand{\arraystretch}{1.3}
\setlength{\tabcolsep}{3.5pt}
\begin{tabular}{cccccc}
\toprule
\makecell[c]{Algorithm} &
\makecell[c]{Epochs} &
\makecell[c]{Training/\\Testing} &
\makecell[c]{Accuracy\\(Exp. data)} &
\makecell[c]{Average\\ accuracy\\(Exp. data)} &
\makecell[c]{Accuracy\\(Sim. data)} \\
\midrule

\multirow{3}{*}{ResNet-50} &
\multirow{3}{*}{30} &
8:2 & 90.47\% & \multirow{3}{*}{90.28\%} & 97.89\% \\
& & 7:3 & 89.95\% & & 97.41\% \\
& & 6:4 & 90.41\% & & 97.74\% \\
\midrule

\multirow{3}{*}{ResNet-34} &
\multirow{3}{*}{20} &
8:2 & 90.19\% & \multirow{3}{*}{90.22\%} & 97.61\% \\
& & 7:3 & 90.07\% & & 97.26\% \\
& & 6:4 & 90.39\% & & 97.82\% \\
\midrule

\multirow{3}{*}{ResNet-18} &
\multirow{3}{*}{20} &
8:2 & 90.42\% & \multirow{3}{*}{90.17\%} & 97.80\% \\
& & 7:3 & 90.13\% & & 97.64\% \\
& & 6:4 & 89.96\% & & 97.72\% \\
\midrule

\multirow{3}{*}{VGG-19} &
\multirow{3}{*}{20} &
8:2 & 89.75\% & \multirow{3}{*}{89.99\%} & 97.77\% \\
& & 7:3 & 90.26\% & & 97.81\% \\
& & 6:4 & 89.96\% & & 97.39\% \\
\bottomrule

\end{tabular}
\end{table}

 In the training process, fusion reaction events are labeled as 1 (positive samples), and elastic scattering events are labeled as 0 (negative samples). For the classification results of ResNet-50 on experimental data, the numbers of true positives (TP), false positives (FP), false negatives (FN), and true negatives (TN), as obtained from the confusion matrix, are shown in Fig. \ref{fig5} (c). To evaluate the performance of the model, several evaluation metrics are calculated, with Precision = 0.948, Recall = 0.914, F1 score = 0.931, and False Positive Rate (FPR) = 0.117. The high Precision indicates that most predicted positive events are correct, while the strong Recall shows that most true positive events are detected. The F1 Score reflects a balanced performance between Precision and Recall, capturing both the accuracy of positive predictions and the ability to detect true positive events. The low FPR indicates that the model performs few incorrect classifications, further demonstrating its robustness and reliability.

\subsubsection{Analysis of the mislabeled events}\label{sec8}
\begin{figure*}[htbp]
    \centering

    \setlength{\abovecaptionskip}{1pt}  
    \setlength{\belowcaptionskip}{0pt}  
    \setlength{\subfigbottomskip}{0pt}  
    \setlength{\subfigcapskip}{0pt}

    \subfigure{
        \includegraphics[width=0.45\textwidth]{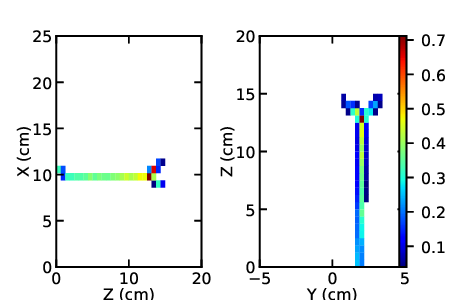}
    }
    \subfigure{
        \includegraphics[width=0.45\textwidth]{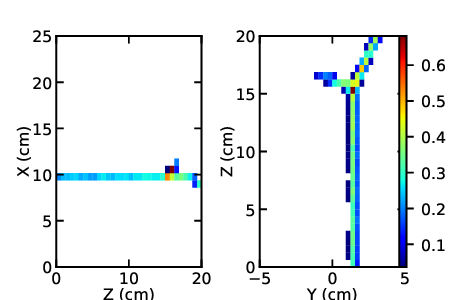}
    }

    \caption{Two elastic scattering events in the experimental data that are misclassified as fusion reaction events by traditional methods, but machine learning distinguishes the two events as elastic scattering events.}
    \label{fig7}
\end{figure*}

We analyze the experimental data that are misclassified by the ResNet-50 model and find that some of these cases are in fact mislabeled by traditional methods. For these discrepant events, a further event-by-event manual inspection combined with track visualization is performed. About $1.5\%$ of the experimental data are mislabeled by the traditional methods. Interestingly, the ResNet-50 model successfully identifies certain misclassified events. For instance, due to the similarity in track patterns between elastic scattering events and fusion reaction events, some elastic scattering events are mislabeled as fusion reactions by traditional methods. As shown in Fig.~\ref{fig7}, the two events are elastic scattering events, but they are labeled as fusion reaction events by traditional methods. However, the ResNet-50 model correctly identified them as elastic scattering events. 

\begin{figure*}[htbp]
    \centering

    \setlength{\abovecaptionskip}{1pt}  
    \setlength{\belowcaptionskip}{0pt}  
    \setlength{\subfigbottomskip}{0pt}  
    \setlength{\subfigcapskip}{0pt}

    \subfigure{
        \includegraphics[width=0.45\textwidth,height=5.5cm]{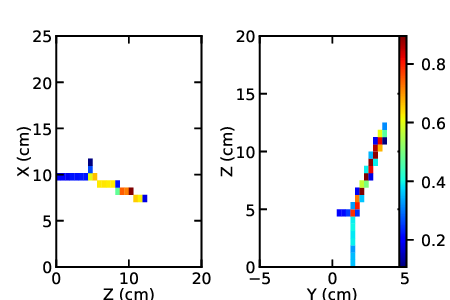}
    }
    \subfigure{
        \includegraphics[width=0.45\textwidth,height=5.5cm]{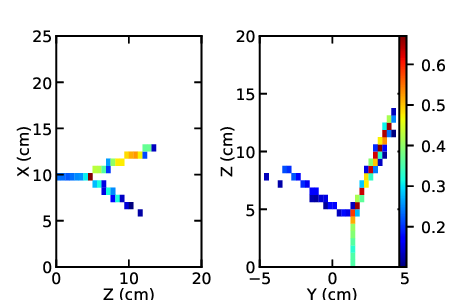}
    }

    \caption{Two fusion reaction events in the experimental data that are misclassified by the machine learning.}
    \label{fig8}
\end{figure*}

Additionally, some misclassifications are caused by false positives from the machine learning model, as shown in Fig.~\ref{fig8}. This is mainly due to the fact that the simulated data cannot perfectly match the experimental data, resulting in false predictions by the model. 

These results demonstrate that machine learning can complement traditional data analysis techniques, achieving improved accuracy in identifying elastic scattering and fusion reaction events, and thereby reducing the false positive rate.

\subsubsection{Classification of events from different fusion reaction channels}\label{sec9}
\begin{figure}[htbp]
    \centering
    \setlength{\abovecaptionskip}{1pt}  
    \setlength{\belowcaptionskip}{0pt}  
    \setlength{\subfigbottomskip}{0pt}  
    \setlength{\subfigcapskip}{0pt}

    \includegraphics[width=0.45\textwidth]{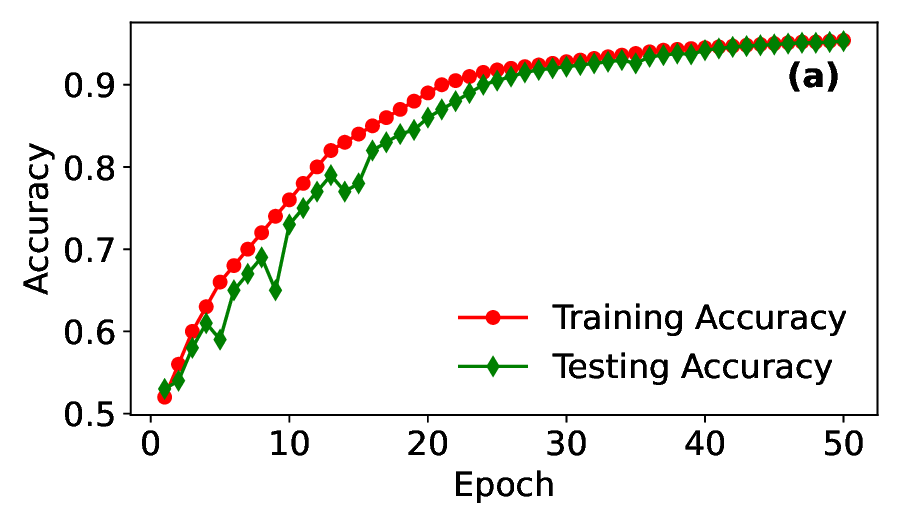}\\[-3pt]
    \includegraphics[width=0.45\textwidth]{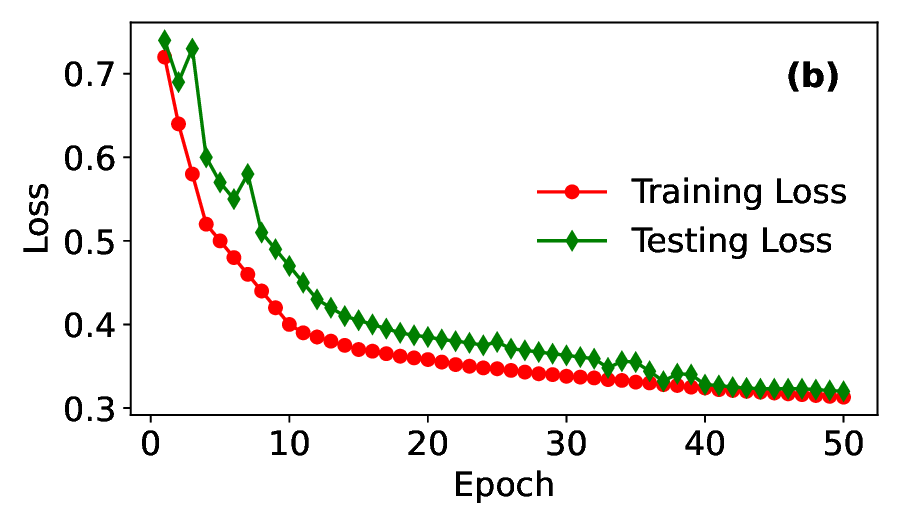}\\[-3pt]
    \includegraphics[width=0.47\textwidth]{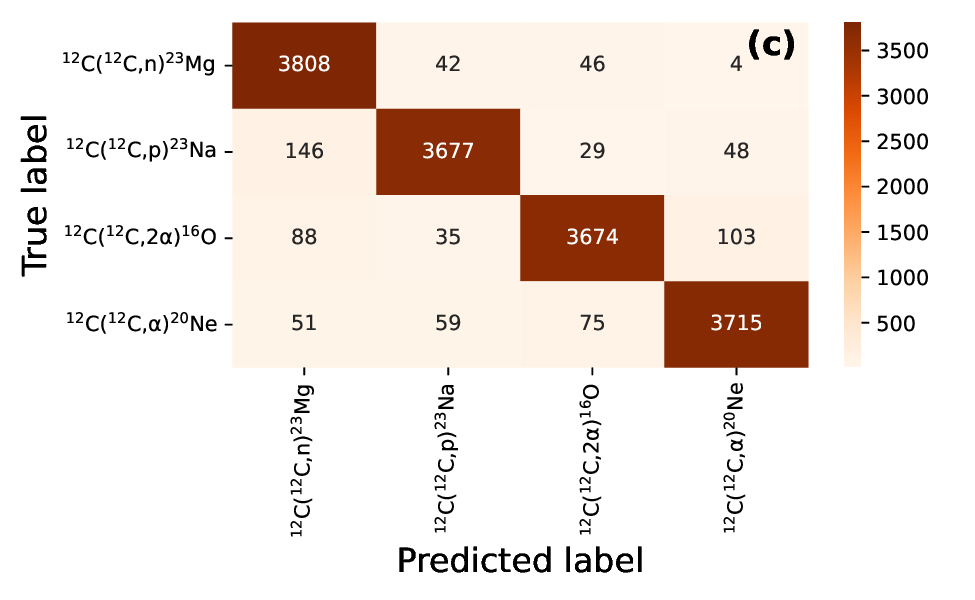}
    
    \caption{
The accuracy (a) and loss (b) of the training and testing sets over epoch for the classification of simulated fusion reaction events using ResNet-50. (c) Confusion matrix obtained from classifying events of different fusion reaction channels. The values along the diagonal represent the number of correctly classified events, while the off-diagonal values indicate the number of misclassified events.}
    \label{fig9}
\end{figure}

The major channels of the fusion reaction include $^{12}\mathrm{C}(^{12}\mathrm{C}, p)^{23}\mathrm{Na}$, $^{12}\mathrm{C}(^{12}\mathrm{C}, \alpha)^{20}\mathrm{Ne}$, $^{12}\mathrm{C}(^{12}\mathrm{C}, n)^{23}\mathrm{Mg}$ and $^{12}\mathrm{C}(^{12}\mathrm{C}, 2\alpha)^{16}\mathrm{O}$. For the proton and neutron reaction channels, they cannot be distinguished experimentally because the proton cannot be detected. However, in the simulated data, the proton tracks are available, enabling the CNN to distinguish the two reaction channels based on the track features of the proton. To identify specific fusion reaction channels, we train the ResNet-50, ResNet-34, ResNet-18, and VGG-19 models to classify simulated events of different fusion reactions and select events corresponding to a channel of interest. The classification results of the four models are shown in Table \ref{tab3}. The results show that the classification accuracies of the four models are around $95\%$. Figs. \ref{fig9} (a) and (b) show the accuracy and loss of the training and testing sets over epoch, respectively. The confusion matrix of correctly and incorrectly classified events of ResNet-50 is shown in Fig. \ref{fig9} (c). These results demonstrate that ResNet and VGG can effectively distinguish different fusion reaction channels, providing valuable support for further analysis and selection of events from a channel of interest. 

\begin{table}[htbp]
\centering
\caption{Classification results of four CNN models on different fusion reaction channels in simulated data.}
\begin{tabular}{ccc}
\hline
Algorithm & Epochs & Accuracy \\
\hline
ResNet-50 & 50 & $95.35\%$ \\
ResNet-34 & 30 & $95.73\%$ \\
ResNet-18 & 30 & $95.83\%$ \\
VGG-19 & 30 & $94.98\%$ \\
\hline
\end{tabular}
\label{tab3}
\end{table}

\subsubsection{Reconstruction of reaction vertex using CNN}\label{sec10}
Initially, we try to use the ResNet and VGG architectures for the reconstruction of reaction vertex. However, due to the problem of overfitting, effective reconstruction of the reaction vertex cannot be achieved. Therefore, in this study, we designed a CNN model to reconstruct the reaction vertex of the events. 

\begin{figure*}[htbp]
    \centering

    \setlength{\abovecaptionskip}{1pt}  
    \setlength{\belowcaptionskip}{0pt}  
    \setlength{\subfigbottomskip}{0pt}  
    \setlength{\subfigcapskip}{0pt}

    \subfigure{
        \includegraphics[width=0.3\textwidth]{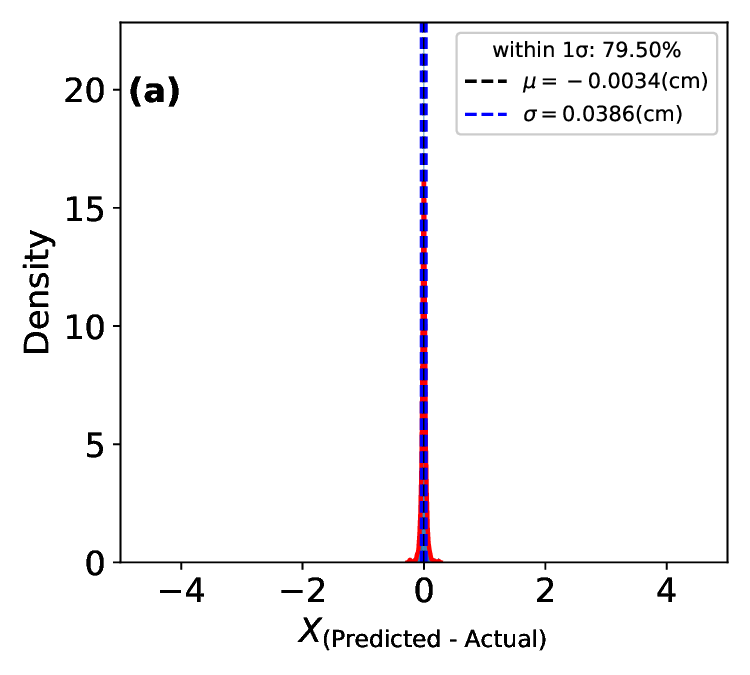}
    }
    \subfigure{
        \includegraphics[width=0.3\textwidth]{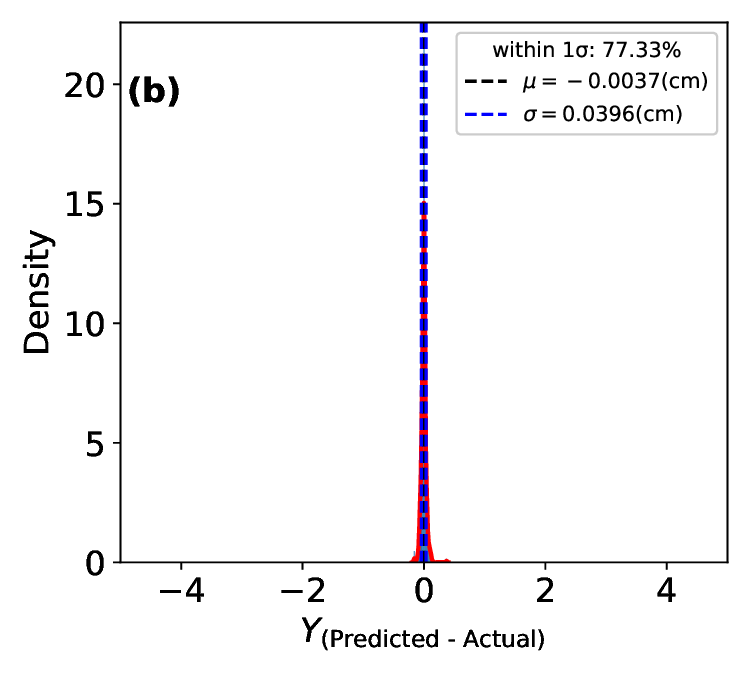}
    }
    \subfigure{
        \includegraphics[width=0.3\textwidth]{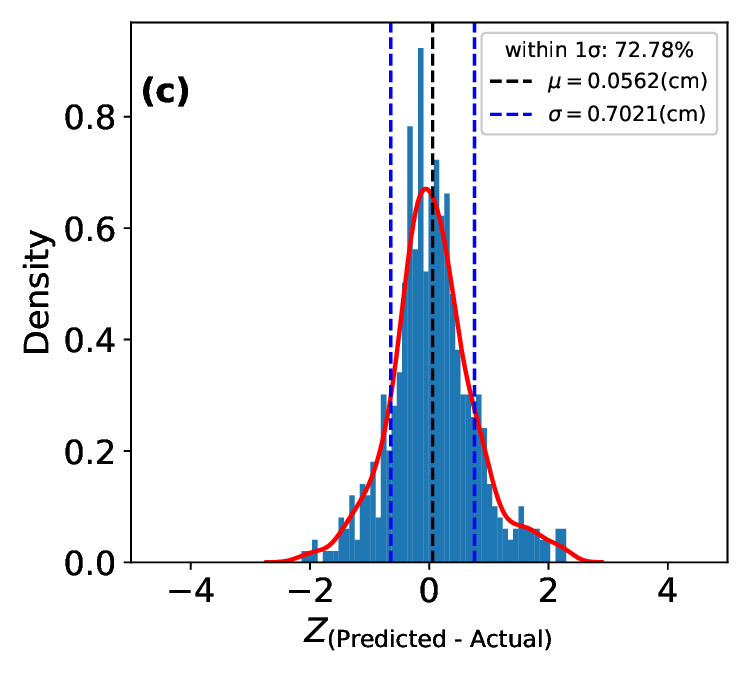}
    }
    \subfigure{
        \includegraphics[width=0.3\textwidth]{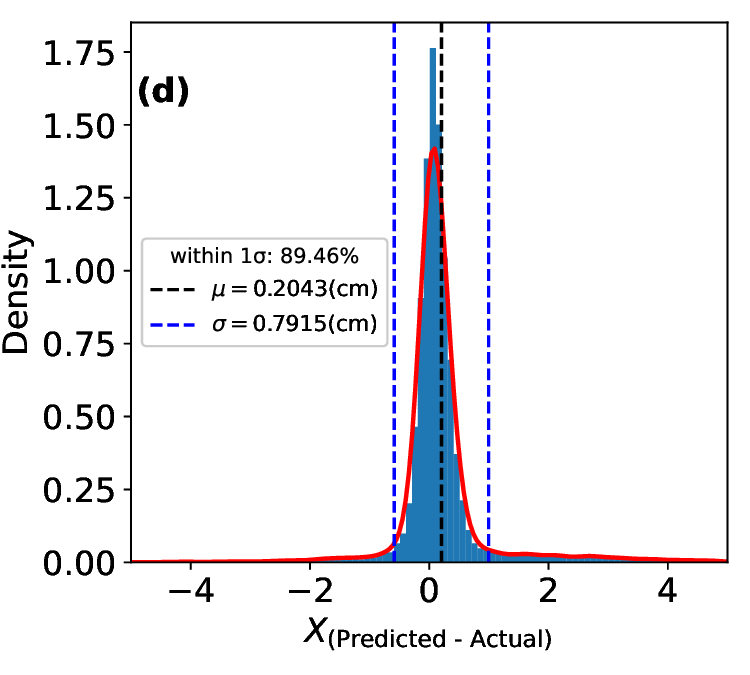}
    }
    \subfigure{
        \includegraphics[width=0.3\textwidth]{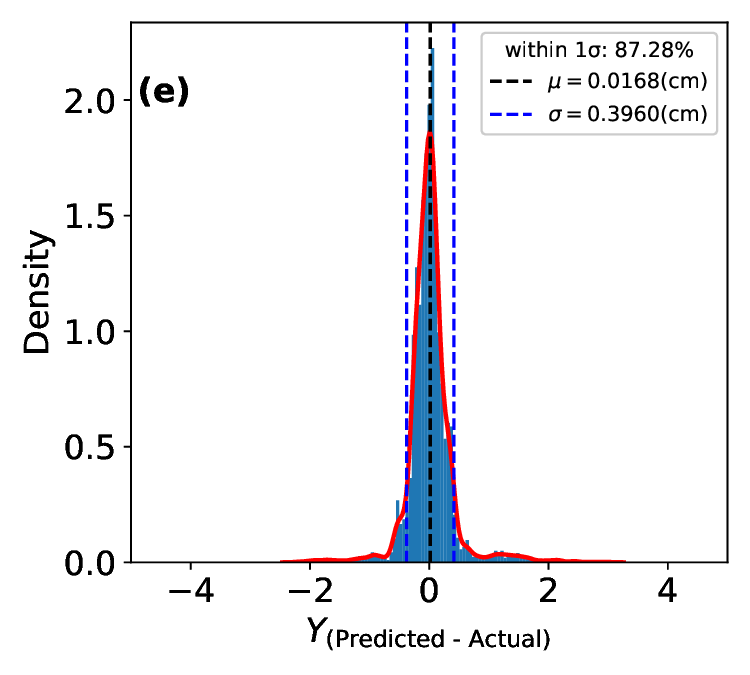}
    }
    \subfigure{
        \includegraphics[width=0.3\textwidth]{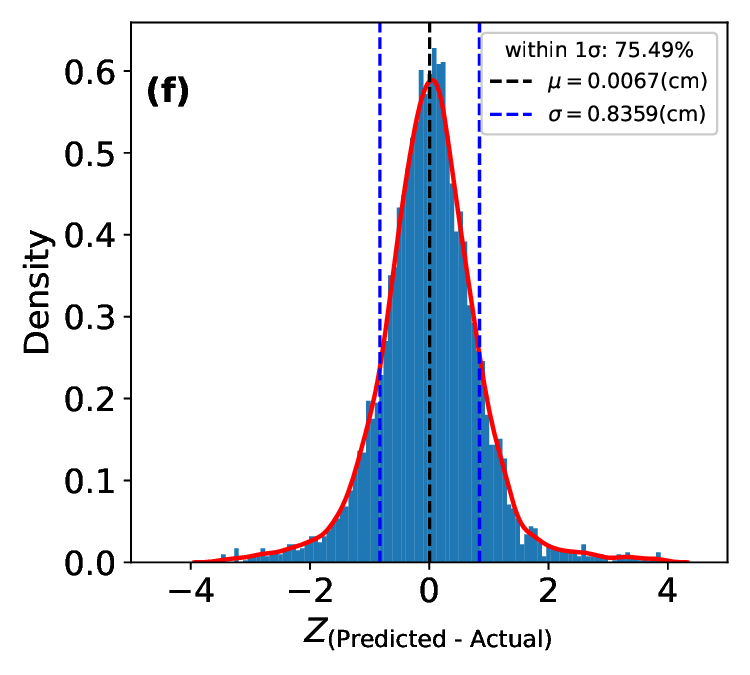}
    }
    
    \caption{The residual distributions between the reconstructed and true values of $\mathrm{vertexX}$ (a), $\mathrm{vertexY}$ (b), and $\mathrm{vertexZ}$ (c) for simulated data using the CNN model. The reconstruction of the simulation-trained network when applied to experimental data for the $\mathrm{vertexX}$ (d), $\mathrm{vertexY}$ (e), and $\mathrm{vertexZ}$ (f). The true vertex values of the experimental data are obtained according to Ref. \cite{Zhang2021}.}
    \label{fig12}
\end{figure*}

The architecture of the CNN model is shown in Table~\ref{tab1}, with the input data represented as 128$\times$128 pixels color images as previously described. The model is trained using the Adam optimizer with a learning rate of 0.001, and the Mean Squared Error (MSE) loss function is employed to measure the error between predicted and true values. We first train the CNN model using simulated data, and the trained network is then applied to reconstruct the reaction vertex from experimental data. Figs. \ref{fig12} (a)-(c) present the residual distributions between the reconstructed and true vertex values for the simulated data, with standard deviations of residuals for vertex X, Y, and Z being \(0.0386~\mathrm{cm}\), \(0.0396~\mathrm{cm}\), and \(0.7021~\mathrm{cm}\), respectively. Figs. (d)-(f) show the corresponding results for experimental data obtained with the simulation-trained network, where the standard deviations are \(0.7915~\mathrm{cm}\), \(0.3960~\mathrm{cm}\), and \(0.8359~\mathrm{cm}\), respectively. 

\begin{figure*}[htbp]
    \centering

    \setlength{\abovecaptionskip}{1pt}  
    \setlength{\belowcaptionskip}{0pt}  
    \setlength{\subfigbottomskip}{0pt}  
    \setlength{\subfigcapskip}{0pt}

    \subfigure{
        \includegraphics[width=0.3\textwidth]{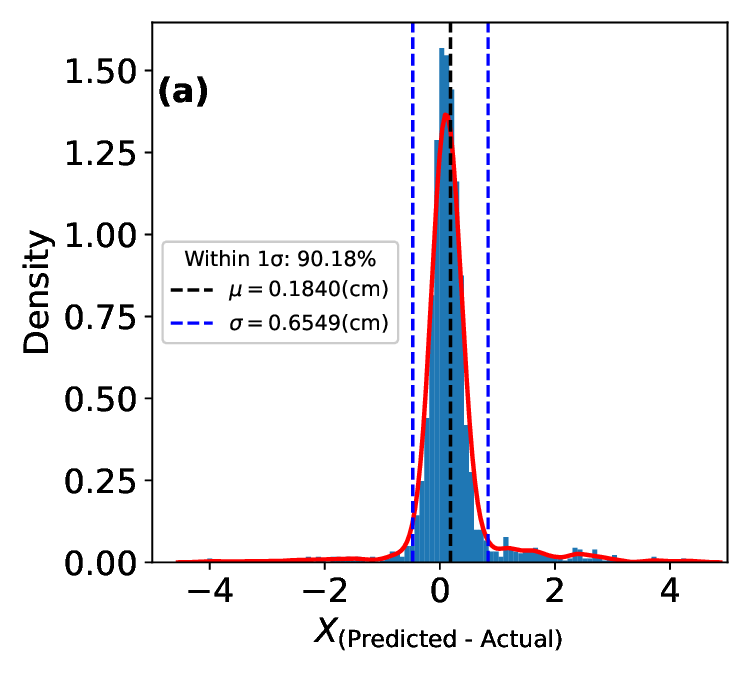}
    }
    \subfigure{
        \includegraphics[width=0.3\textwidth]{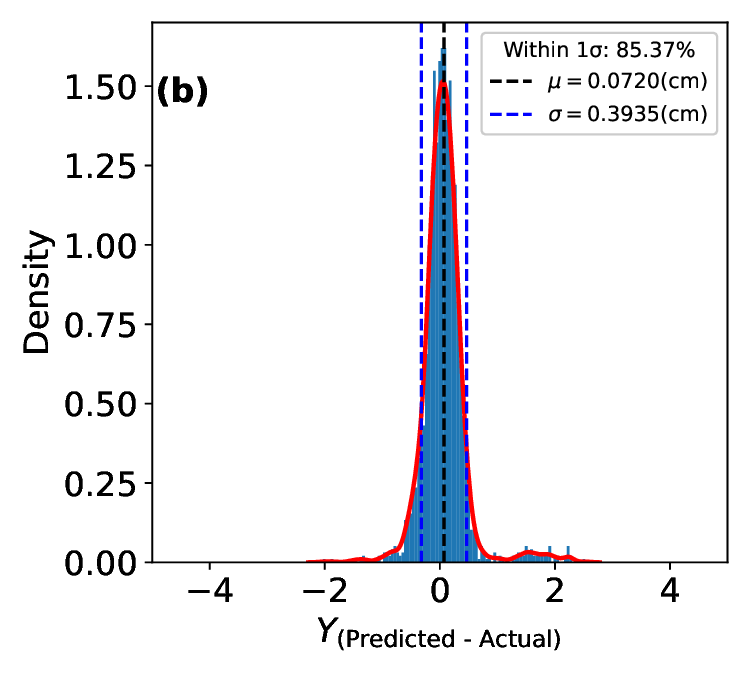}
    }
    \subfigure{
        \includegraphics[width=0.3\textwidth]{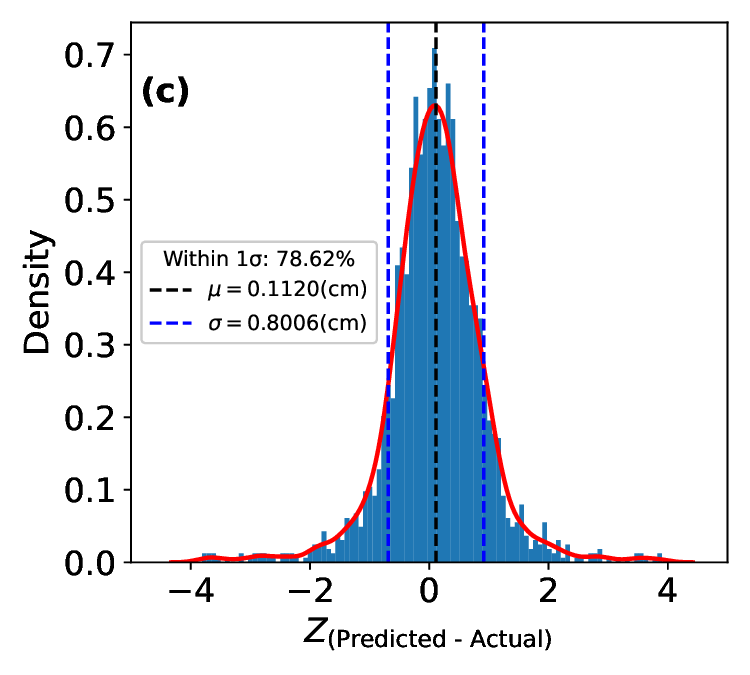}
    }
    
    \caption{The residual distributions between the true and predicted values of $\mathrm{vertexX}$ (a), $\mathrm{vertexY}$ (b), and $\mathrm{vertexZ}$ (c) obtained using the CNN model trained on the experimental data.}
    \label{fig11}
\end{figure*}

In addition, the CNN is trained directly on experimental data to examine its reconstruction performance. The results are displayed in Fig. \ref{fig11}, yielding standard deviations of \(0.6549~\mathrm{cm}\), \(0.3935~\mathrm{cm}\), and \(0.8006~\mathrm{cm}\), respectively. A comparison between Fig. \ref{fig11} and Fig. \ref{fig12} indicates that the reconstruction of the reaction vertex from experimental data using the simulation-trained network is largely consistent with that obtained from the network trained directly on experimental data. 

 Within the low-energy regime, fusion reactions and small-angle elastic scattering events exhibit highly similar track geometries and energy-loss distributions. Consequently, the vertex-reconstruction accuracy is primarily limited by experimental geometry and detector-related factors, rather than by differences among reaction channels.

The poorer vertex reconstruction resolution along the Z direction compared to the Y direction mainly arises from the intrinsic spatial resolution limitation imposed by the geometry of the readout plane. The readout plane is segmented into $32\times32$ rectangular pads, each with a size of 0.3 cm(Y)$\times$0.6 cm(Z). This asymmetric pad geometry reduces position precision along Z, resulting in a larger uncertainty in the reconstructed vertex.

\begin{table}[htbp]
\centering
\caption{The Structure of CNN Model.}
\begin{tabular}{lll}
\hline
Layer & Type & Output Shape \\
\hline
Input Layer & Input & (128, 128, 3) \\
Conv2D & 32 filters, 3$\times$3, ReLU & (128, 128, 32) \\
Max Pooling2D & 2$\times$2 & (64, 64, 32) \\
Conv2D & 64 filters, 3$\times$3, ReLU & (64, 64, 64) \\
Max Pooling2D & 2$\times$2 & (32, 32, 64) \\
Conv2D & 128 filters, 3$\times$3, ReLU & (32, 32, 128) \\
Global Average Pooling2D & - & (128) \\
Fully Connected & 128 units, ReLU & (128) \\
Dropout & rate=0.3 & (128) \\
Output Layer & Dense (3) & (3) \\
\hline
\end{tabular}
\label{tab1}
\end{table}

\section{Conclusion and Discussion}\label{sec11}
In this work, we successfully applied machine learning techniques to analyze experimental data of the ${}^{12}\mathrm{C} + {}^{12}\mathrm{C}$ fusion reaction around the Coulomb barrier. The dataset contains a large number of elastic scattering events mixed with fusion reaction events, which significantly complicates the determination of fusion reaction cross section. We first train the ResNet-50, ResNet-34, ResNet-18, and VGG-19 models on simulated data to classify elastic scattering and fusion reaction events. The performance of the four models is evaluated on experimental data, showing the accuracies of approximately $97\%$ for the simulated data and $90\%$ for the experimental data. Through analysis of misclassified events, we find that machine learning can correctly label some events that are misclassified by traditional methods. To identify specific fusion reaction channels, we train the four models to classify simulated events from different fusion reactions channels to select the events corresponding to a channel of interest. The results show that the four models achieve the classification accuracies are approximately $95\%$. Furthermore, we developed a CNN model to reconstruct the reaction vertex. The results show that the fraction of residuals falling within the 1$\sigma$ range is approximately $80\%$, confirming that the CNN can effectively reconstruct the reaction vertex.

 In the future, we aim to employ machine learning to establish an automated and intelligent data processing pipeline, to overcome the limitations of conventional approaches and enhancing both the efficiency and accuracy of nuclear reaction data analysis. With Active Target time projection chamber evolving toward higher counting rates, the resulting data volume is expected to increase substantially, rendering traditional analysis methods insufficient to handle such a data deluge. The intelligent analysis framework developed in this work is conceived as a forward-looking solution, designed to fully exploit the potential of next-generation detectors and to provide the technical foundation for processing larger and more complex datasets. Machine learning is expected to be widely used for the analysis of nuclear reaction data.

However, machine learning analysis methods exhibit inherent limitations, such as relatively high false-positive rates. This is possibly due to the fact that simulated data cannot be perfectly aligned with experimental data. These factors can affect both the precision and accuracy of data analysis. Consequently, continuous improvement of neural network architectures and optimization of both simulated and experimental data are essential for advancing future development.



\begin{thebibliography}{99}

\bibitem{Blumenfeld2013}
Y. Blumenfeld, T. Nilsson, P. Van Duppen,
Facilities and methods for radioactive ion beam production.
Phys. Scr. {\bf T152}, 014023 (2013).
\href{https://doi.org/10.1088/0031-8949/2013/T152/014023}{doi:10.1088/0031-8949/2013/T152/014023}

\bibitem{Stracener2003}
D.W. Stracener,
Status of radioactive ion beams at the HRIBF.
Nucl. Instrum. Meth. B {\bf 204}, 42--47 (2003).
\href{https://doi.org/10.1016/S0168-583X(02)01888-8}{doi:10.1016/S0168-583X(02)01888-8}

\bibitem{Bricault1997}
P.G. Bricault, M. Dombsky, P.W. Schmor et al.,
Radioactive ion beams facility at TRIUMF.
Nucl. Instrum. Meth. B {\bf 126}, 231--235 (1997).
\href{https://doi.org/10.1016/S0168-583X(96)01037-3}{doi:10.1016/S0168-583X(96)01037-3}

\bibitem{bazin2020low}
D. Bazin, T. Ahn, Y. Ayyad et al.,
Low energy nuclear physics with active targets and time projection chambers.
Prog. Part. Nucl. Phys. {\bf 114}, 103790 (2020).
\href{https://doi.org/10.1016/j.ppnp.2020.103790}{doi:10.1016/j.ppnp.2020.103790}

\bibitem{ayyad2018physics}
Y. Ayyad, D. Bazin, S. Beceiro-Novo et al.,
Physics and technology of time projection chambers as active targets.
Eur. Phys. J. A {\bf 54}, 181 (2018).
\href{https://doi.org/10.1140/epja/i2018-12557-7}{doi:10.1140/epja/i2018-12557-7}

\bibitem{MITTIG2015494}
W. Mittig, S. Beceiro-Novo, A. Fritsch et al., Active Target detectors for studies with exotic beams: Present and next future. 
Nucl. Instrum. Meth. A {\bf 784}, 494--498 (2015). 
\href{https://doi.org/10.1016/j.nima.2014.10.048}{doi:10.1016/j.nima.2014.10.048}

\bibitem{GIOVINAZZO2020}
J. Giovinazzo, J. Pancin, J. Pibernat et al.,
ACTAR TPC performance with GET electronics.
Nucl. Instrum. Meth. A {\bf 953}, 163184 (2020).
\href{https://doi.org/10.1016/j.nima.2019.163184}{doi:10.1016/j.nima.2019.163184}

\bibitem{KOSHCHIY2020}
E. Koshchiy, G.V. Rogachev, E. Pollacco et al.,
Texas Active Target (TexAT) detector for experiments with rare isotope beams.
Nucl. Instrum. Meth. A {\bf 957}, 163398 (2020).
\href{https://doi.org/10.1016/j.nima.2020.163398}{doi:10.1016/j.nima.2020.163398}

\bibitem{OBERLA2016}
E. Oberla, H.J. Frisch,
The design and performance of a prototype water Cherenkov optical time-projection chamber.
Nucl. Instrum. Meth. A {\bf 814}, 19--32 (2016).
\href{https://doi.org/10.1016/j.nima.2016.01.030}{doi:10.1016/j.nima.2016.01.030}

\bibitem{WU2023168528}
H.K. Wu, Y.J. Wang, Y.M. Wang et al., Machine learning method for $^{12}$C event classification and reconstruction in the active target time-projection chamber. 
Nucl. Instrum. Meth. A {\bf 1055}, 168528 (2023). 
\href{https://doi.org/10.1016/j.nima.2023.168528}{doi:10.1016/j.nima.2023.168528}

\bibitem{ZHANG2021165740}
Z.C. Zhang, X.Y. Wang, T.L. Pu et al., Studying the heavy-ion fusion reactions at stellar energies using Time Projection Chamber. 
Nucl. Instrum. Meth. A {\bf 1016}, 165740 (2021). 
\href{https://doi.org/10.1016/j.nima.2021.165740}{doi:10.1016/j.nima.2021.165740}

\bibitem{Li2024}
X.B. Li, L.H. Ru, Z.C. Zhang et al.,
Construction and performance test of charged particle detector array for MATE.
Nucl. Sci. Tech. {\bf 35}, 131 (2024).
\href{https://doi.org/10.1007/s41365-024-01500-7}{doi:10.1007/s41365-024-01500-7}

\bibitem{LI2024}
Y. Li, Y. Han, Y.K. Sun et al.,
Performance study of the Multi-purpose Time Projection Chamber (MTPC) using a four-component alpha source.
Nucl. Instrum. Meth. A {\bf 1060}, 169045 (2024).
\href{https://doi.org/10.1016/j.nima.2023.169045}{doi:10.1016/j.nima.2023.169045}

\bibitem{Chen2025}
J. Chen, Y. Ayyad, D. Bazin et al.,
Near-Threshold Dipole Strength in $^{10}\mathrm{Be}$ with Isoscalar Character.
Phys. Rev. Lett. {\bf 134}, 012502 (2025).
\href{https://doi.org/10.1103/PhysRevLett.134.012502}{doi:10.1103/PhysRevLett.134.012502}

\bibitem{Chen2024}
J. Chen, J.R. Ma,
Inelastic scattering reaction as a probe for monopole, dipole and quadrupole excitations.
EPJ Web Conf. {\bf 311}, 00008 (2024).
\href{https://doi.org/10.1051/epjconf/202431100008}{doi:10.1051/epjconf/202431100008}

\bibitem{Zhang2023prc}
S. Zhang, G. Li, W. Jiang et al., 
Measurement of the $^{159}$Tb(n,$\gamma$) cross section at the CSNS Back-n facility.
Phys. Rev. C {\bf 107}, 045809 (2023).
\href{https://doi.org/10.1103/PhysRevC.107.045809}{doi:10.1103/PhysRevC.107.045809}

\bibitem{RevModPhys.94.031003}
A. Boehnlein, M. Diefenthaler, N. Sato et al., Colloquium: Machine learning in nuclear physics. 
Rev. Mod. Phys. {\bf 94}, 031003 (2022). 
\href{https://doi.org/10.1103/RevModPhys.94.031003}{doi:10.1103/RevModPhys.94.031003}

\bibitem{he2023high}
W.B. He, Y.G. Ma, L.G. Pang et al.,
High-energy nuclear physics meets machine learning.
Nucl. Sci. Tech. {\bf 34}, 88 (2023).
\href{https://doi.org/10.1007/s41365-023-01233-z}{doi:10.1007/s41365-023-01233-z}

\bibitem{GAO2021}
Z.P. Gao, Y.J. Wang, H.L. Lü et al.,
Machine learning the nuclear mass.
Nucl. Sci. Tech. {\bf 32}, 109 (2021).
\href{https://doi.org/10.1007/s41365-021-00956-1}{doi:10.1007/s41365-021-00956-1}

\bibitem{YUAN2024}
Z.Y. Yuan, D. Bai, Z. Wang et al.,
Reliable calculations of nuclear binding energies by the Gaussian process of machine learning.
Nucl. Sci. Tech. {\bf 35}, 105 (2024).
\href{https://doi.org/10.1007/s41365-024-01463-9}{doi:10.1007/s41365-024-01463-9}

\bibitem{SHANG2022}
T.S. Shang, J. Li, Z.M. Niu et al.,
Prediction of nuclear charge density distribution with feedback neural network.
Nucl. Sci. Tech. {\bf 33}, 153 (2022).
\href{https://doi.org/10.1007/s41365-022-01140-9}{doi:10.1007/s41365-022-01140-9}

\bibitem{HE2021}
J. He, W.B. He, Y.G. Ma et al.,
Machine-learning-based identification for initial clustering structure in relativistic heavy-ion collisions.
Phys. Rev. C {\bf 104}, 044902 (2021).
\href{https://doi.org/10.1103/PhysRevC.104.044902}{doi:10.1103/PhysRevC.104.044902}

\bibitem{KIM2023}
C.H. Kim, S. Ahn, K.Y. Chae et al.,
Noise signal identification in time projection chamber data using deep learning model.
Nucl. Instrum. Meth. A {\bf 1048}, 168025 (2023).
\href{https://doi.org/10.1016/j.nima.2023.168025}{doi:10.1016/j.nima.2023.168025}

\bibitem{QIAN2021}
Z. Qian, V. Belavin, V. Bokov et al.,
Vertex and energy reconstruction in JUNO with machine learning methods.
Nucl. Instrum. Meth. A {\bf 1010}, 165527 (2021).
\href{https://doi.org/10.1016/j.nima.2021.165527}{doi:10.1016/j.nima.2021.165527}

\bibitem{MAYER2021}
J. Mayer, K. Boretzky, C. Douma et al.,
Classical and machine learning methods for event reconstruction in NeuLAND.
Nucl. Instrum. Meth. A {\bf 1013}, 165666 (2021).
\href{https://doi.org/10.1016/j.nima.2021.165666}{doi:10.1016/j.nima.2021.165666}

\bibitem{LI2022}
Z.Y. Li, Z. Qian, J.H. He et al.,
Improvement of machine learning-based vertex reconstruction for large liquid scintillator detectors with multiple types of PMTs.
Nucl. Sci. Tech. {\bf 33}, 93 (2022).
\href{https://doi.org/10.1007/s41365-022-01078-y}{doi:10.1007/s41365-022-01078-y}

\bibitem{DELAQUIS2018}
S. Delaquis, M.J. Jewell, I. Ostrovskiy et al.,
Deep neural networks for energy and position reconstruction in EXO-200.
J. Instrum. {\bf 13}, P08023 (2018).
\href{https://doi.org/10.1088/1748-0221/13/08/P08023}{doi:10.1088/1748-0221/13/08/P08023}

\bibitem{KUCHERA2019156}
M.P. Kuchera, R. Ramanujan, J.Z. Taylor et al., Machine learning methods for track classification in the AT-TPC. 
Nucl. Instrum. Meth. A {\bf 940}, 156--167 (2019). 
\href{https://doi.org/10.1016/j.nima.2019.05.097}{doi:10.1016/j.nima.2019.05.097}

\bibitem{SOLLI2021165461}
R. Solli, D. Bazin, M. Hjorth-Jensen et al., Unsupervised learning for identifying events in active target experiments. 
Nucl. Instrum. Meth. A {\bf 1010}, 165461 (2021). 
\href{https://doi.org/10.1016/j.nima.2021.165461}{doi:10.1016/j.nima.2021.165461}

\bibitem{GHIMIRE2025165649}
R. Ghimire, A. Ratkiewicz, S.D. Pain et al., Background subtraction in inelastic scattering measurements using machine learning. 
Nucl. Instrum. Meth. B {\bf 561}, 165649 (2025). 
\href{https://doi.org/10.1016/j.nimb.2025.165649}{doi:10.1016/j.nimb.2025.165649}

\bibitem{DEY2025170002}
P. Dey, A.K. Anthony, C. Hunt et al., Point-cloud based machine learning for classifying rare events in the Active-Target Time Projection Chamber. 
Nucl. Instrum. Meth. A {\bf 1072}, 170002 (2025). 
\href{https://doi.org/10.1016/j.nima.2024.170002}{doi:10.1016/j.nima.2024.170002}

\bibitem{MATEroot2025}
L. Li, Z.C. Zhang, N.T. Zhang et al.,
MATEROOT: A Simulation and Analysis Tool for Experiments with MATE.
ChinaXiv:202605.00027.
\href{https://doi.org/10.12074/202605.00027}{doi:10.12074/202605.00027}

\bibitem{XIA200211}
J.W. Xia, W.L. Zhan, B.W. Wei et al., The heavy ion cooler-storage-ring project (HIRFL-CSR) at Lanzhou. 
Nucl. Instrum. Meth. A {\bf 488}, 11--25 (2002). 
\href{https://doi.org/10.1016/S0168-9002(02)00475-8}{doi:10.1016/S0168-9002(02)00475-8}

\bibitem{Nan2024}
W.K. Nan, Y.B. Wang, Y.D. Sheng et al.,
Novel thick-target inverse kinematics method for the astrophysical $^{12}\mathrm{C}+\,^{12}\mathrm{C}$ fusion reaction.
Nucl. Sci. Tech. {\bf 35}, 208 (2024).
\href{https://doi.org/10.1007/s41365-024-01573-4}{doi:10.1007/s41365-024-01573-4}

\bibitem{tang201912c+}
X.D. Tang, L.H. Ru, The $^{12}$C+$^{12}$C fusion reaction at stellar energies. EPJ Web Conf. {\bf 260}, 01002 (2022).
\href{https://doi.org/10.1051/epjconf/202226001002}{doi:10.1051/epjconf/202226001002}

\bibitem{RevModPhys.86.317}
B.B. Back, H. Esbensen, C.L. Jiang et al., Recent developments in heavy-ion fusion reactions. 
Rev. Mod. Phys. {\bf 86}, 317--360 (2014). 
\href{https://doi.org/10.1103/RevModPhys.86.317}{doi:10.1103/RevModPhys.86.317}

\bibitem{PhysRevC.76.035802}
L.R. Gasques, E.F. Brown, A. Chieffi et al., Implications of low-energy fusion hindrance on stellar burning and nucleosynthesis. 
Phys. Rev. C {\bf 76}, 035802 (2007). 
\href{https://doi.org/10.1103/PhysRevC.76.035802}{doi:10.1103/PhysRevC.76.035802}

\bibitem{Wang20251}
S. Wang, Y.Z. Li, L.H. Ru et al.,
$^{12}$C+$^{12}$C fusion reaction at astrophysical energies using HOPG target.
Nucl. Sci. Tech. {\bf 36}, 143 (2025).
\href{https://doi.org/10.1007/s41365-025-01714-3}{doi:10.1007/s41365-025-01714-3}

\bibitem{Tang20191}
X.D. Tang, S.B. Ma, X. Fang et al.,
An efficient method for mapping the $^{12}$C+$^{12}$C molecular resonances at low energies.
Nucl. Sci. Tech. {\bf 30}, 126 (2019).
\href{https://doi.org/10.1007/s41365-019-0652-9}{doi:10.1007/s41365-019-0652-9}

\bibitem{Liu20241}
W.P. Liu, B. Guo, Z. An et al.,
Recent progress in nuclear astrophysics research and its astrophysical implications at the China Institute of Atomic Energy.
Nucl. Sci. Tech. {\bf 35}, 217 (2024).
\href{https://doi.org/10.1007/s41365-024-01590-3}{doi:10.1007/s41365-024-01590-3}


\bibitem{Zhang2021}
Z.C. Zhang,  
Development and Applications of the Time Projection Chamber for the Cross-section Measurements of the Important Fusion Reactions in Astrophysics,  
Ph.D. thesis, University of Chinese Academy of Sciences (Institute of Modern Physics, CAS), 2021.

\bibitem{Wang_2022}
X.Y. Wang, N.T. Zhang, Z.C. Zhang et al., Studies of the $2\alpha$ and $3\alpha$ channels of the $^{12}$C+$^{12}$C reaction in the range of $E_{\mathrm{c.m.}}=8.9$ to 21 MeV using the active target Time Projection Chamber. 
Chin. Phys. C {\bf 46}, 104001 (2022). 
\href{https://doi.org/10.1088/1674-1137/ac7a1d}{doi:10.1088/1674-1137/ac7a1d}

\bibitem{Koonce2021}
B. Koonce,  
ResNet-50 Convolutional Neural Networks with Swift for TensorFlow: Image Recognition and Dataset Categorization. 
pp.~63--72 (2021).

\bibitem{Liang2020}
J.H. Liang,
Image classification based on RESNET.
J. Phys.: Conf. Ser. {\bf 1634}, 012110 (2020).
\href{https://doi.org/10.1088/1742-6596/1634/1/012110}{doi:10.1088/1742-6596/1634/1/012110}

\bibitem{Li2018}
B.Q. Li, Y.Y. He, An improved ResNet based on the adjustable shortcut connections. 
IEEE Access {\bf 6}, 18967 (2018). 
\href{https://doi.org/10.1109/ACCESS.2018.2814605}{doi:10.1109/ACCESS.2018.2814605}

\bibitem{Sengupta2019}
A. Sengupta, Y. Ye, R. Wang et al.,
Going Deeper in Spiking Neural Networks: VGG and Residual Architectures.
Front. Neurosci. {\bf 13}, 95 (2019).
\href{https://doi.org/10.3389/fnins.2019.00095}{doi:10.3389/fnins.2019.00095}

\bibitem{Tammina2019}
S. Tammina,
Transfer learning using VGG-16 with deep convolutional neural network for classifying images.
Int. J. Sci. Res. Publ. {\bf 9}, 143 (2019).
\href{http://dx.doi.org/10.29322/IJSRP.9.10.2019.p9420 }{doi:10.29322/IJSRP.9.10.2019.p9420}


\end{thebibliography}
\end{document}